\documentclass[10pt,twocolumn,letterpaper]{article}

\usepackage{iccv}
\usepackage{times}
\usepackage{epsfig}
\usepackage{epstopdf}
\usepackage{amsmath}
\usepackage{amssymb}
\usepackage{times}
\usepackage{graphicx}
\usepackage{subfigure}
\usepackage{array}
\usepackage{tabularx}
\usepackage{booktabs}
\usepackage{epstopdf}
\usepackage{dcolumn}
\usepackage{color, colortbl}
\newcolumntype{s}{D{.}{.}{1.2}}
\newcolumntype{d}{D{.}{.}{2.1}}
\newcommand{\ts}[1]{^{[#1]}}

\def\MARK{black}

\definecolor{Gray}{gray}{0.9}

\usepackage[breaklinks=true,bookmarks=false]{hyperref}

\iccvfinalcopy 

\def\iccvPaperID{0393} 

\ificcvfinal\fi
\begin{document}

\title{An Empirical Study of Language CNN for Image Captioning}

\author{Jiuxiang Gu$^{1}$, Gang Wang$^2$, Jianfei Cai$^{3}$, Tsuhan Chen$^{3}$\\
$^1$ ROSE Lab, Interdisciplinary Graduate School, Nanyang Technological University, Singapore \\
$^2$ Alibaba AI Labs, Hangzhou, China \\
$^3$ School of Computer Science and Engineering, Nanyang Technological University, Singapore \\
{\tt\small \{jgu004, asjfcai, tsuhan\}@ntu.edu.sg, gangwang6@gmail.com}
}

\maketitle
\thispagestyle{empty}

\begin{abstract}
Language models based on recurrent neural networks have dominated recent image caption generation tasks.
In this paper, we introduce a language CNN model which is suitable for statistical language modeling tasks and shows competitive performance in image captioning.
In contrast to previous models which predict next word based on one previous word and hidden state, our language CNN is fed with all the previous words and can model the long-range dependencies in history words, which are critical for image captioning.
The effectiveness of our approach is validated on two datasets: Flickr30K and MS COCO.
Our extensive experimental results show that our method outperforms the vanilla recurrent neural network based language models and is competitive with the state-of-the-art methods.
\end{abstract}

\section{Introduction}
Image caption generation is a fundamental problem that involves Computer Vision, Natural Language Processing (NLP), and Machine Learning.
It can be analogous to ``\textit{translating}'' an image to proper sentences.
While this task seems to be easy for human beings, it is quite challenging for machines because it requires the model to understand the image content and express their relationships in a natural language.
Also, the image captioning model should be capable of capturing implicit semantic information of an image and generating humanlike sentences.
As a result, generating accurate captions for an image is not an easy task. 

The recent surge of research interest in image caption generation task is due to the advances in Neural Machine Translation (NMT)~\cite{sutskever2014sequence} and large datasets~\cite{plummer2015flickr30k,lin2014microsoft}.
Most image captioning models follow the encoder-decoder pipeline~\cite{cho2014learning,kulkarni2011baby, mao2014deep,karpathy2015deep, rennie2016self}.
The encoder-decoder framework is recently introduced for sequence-to-sequence learning based on Recurrent Neural Networks~(RNNs) or Long-Short Term Memory~(LSTM) networks.
Both RNNs and LSTM networks can be sequence learners.
However, due to the vanishing gradient problem, RNNs can only remember the previous status for a few time steps.
LSTM network is a special type of RNN architecture designed to solve the vanishing gradient problem in RNNs~\cite{vinyals2015show,jia2015guiding,donahue2015long}.
It introduces a new component called memory cell.
Each memory cell is composed of three gates and a neuron with the self-recurrent connection.
These gates allow the memory cells to keep and access information over a long period of time and make LSTM network capable of learning long-term dependencies.

Although models like LSTM networks have memory cells which aim to memorize history information for long-term, \textcolor{\MARK}{they are still} limited to several time steps because long-term information is gradually diluted at every time step~\cite{weston2014memory}.
Besides, vanilla RNNs-based image captioning models recursively accumulate history information without explicitly modeling the hierarchical structure of word sequences, which clearly have a bottom-up structure~\cite{li2015hierarchical}. 

To better model the hierarchical structure and long-term dependencies in word sequences, in this paper, we adopt a language CNN which applies temporal convolution to extract features from sequences.
Such a method is inspired by works in NLP which have shown CNN is very powerful for text representation~\cite{kalchbrenner2014convolutional,wang2015gen}.
Unlike the vanilla CNN architecture, we drop the pooling operation to keep the relevant information for words representation and investigate the optimum convolutional filters by experiments.
However, only using language CNN fails to model the dynamic temporal behavior.
Hence, we still need to combine language CNN with recurrent networks (\eg, RNN or LSTM).
Our extensive studies show that adding language CNN to a recurrent network helps model sequences consistently and more effectively, and leads to improved results.

To summarize, our primary contribution \textcolor{\MARK}{lies in incorporating} a language CNN, \textcolor{\MARK}{which} is capable of capturing long-range dependencies in sequences, \textcolor{\MARK}{with RNNs for image captioning.}
Our model yields comparable performance with the state-of-the-art approaches on Flickr30k~\cite{plummer2015flickr30k} and MS COCO~\cite{lin2014microsoft}.

\section{Related Works}
The problem of generating natural language descriptions for images has become a hot topic in computer vision community.
Prior to using neural networks for generating descriptions, the classical approach is to pose the problem as a retrieval and ranking problem~\cite{hodosh2013framing, gong2014improving, ordonez2011im2text}.
The main weakness of those retrieval-based approaches is that they \textcolor{\MARK}{cannot} generate proper captions for a new combination of objects.
Inspired by the success of deep neural networks in machine translation~\cite{sutskever2014sequence, cho2014learning, kalchbrenner2013recurrent}, \textcolor{\MARK}{researchers} have proposed to use the encoder-decoder framework for image caption generation~\cite{kiros2014multimodal,mao2014deep,karpathy2015deep,vinyals2015show,donahue2015long, chen2015mind, lebret2015phrase}.
Instead of translating sentences between two languages, the goal of image captioning is to ``\textit{translate}" a query image into a sentence that describes the image.
The earliest approach using neural network for image captioning is proposed by Vinyals~\etal~\cite{vinyals2015show} \textcolor{\MARK}{which is} an encoder-decoder system trained to maximize the log-likelihood of the target image descriptions.
Similarly, Mao~\etal~\cite{mao2014deep} and Donahue~\etal\cite{donahue2015long} use the multimodal fusion layer to fuse the image features and word representation at each time step.
In both cases, \ie, the models in~\cite{mao2014deep} and~\cite{donahue2015long}, the captions are generated from the full images,
\textcolor{\MARK}{while} the image captioning model proposed by Karpathy~\etal~\cite{karpathy2015deep} generates descriptions based on regions.
This work is later followed by Johnson~\etal~\cite{johnson2016densecap} whose method is designed to jointly localize regions and describe each with captions.

Rather than representing an image as a single feature vector from the top-layer of CNNs, some researchers \textcolor{\MARK}{have explored} the structure of networks to explicitly or implicitly model the correlation between \textcolor{\MARK}{images} and descriptions~\cite{xu2015show,lu2016knowing, liu2017attention}.
Xu~\etal~\cite{xu2015show} incorporate the spatial attention on convolutional features of an image into the encoder-decoder framework through the ``\textit{hard}" and ``\textit{soft}" attention mechanisms.
Their work is followed by Yang~\etal~\cite{yang2016encode} whose method introduces a review network to improve the attention mechanism and Liu~\etal~\cite{liu2017attention} whose approach is designed to improve the correctness of visual attention.
Moreover, a variational autoencoder for image captioning is developed by Pu~\etal~\cite{pu2016variational}. They use a CNN as the image encoder and use a deep generative deconvolutional network as the decoder together with a Gated Recurrent Unit (GRU)~\cite{cho2014learning} to generate image descriptions.

More recently, high-level attributes have been shown to obtain clear improvements on \textcolor{\MARK}{the} image captioning task when injected into existing encoder-decoder based models~\cite{wu2016value, you2016image, gan2016learning}.
Specifically, Jia~\etal~\cite{jia2015guiding} use the semantic information as the extra input to guide the model in generating captions.
In addition, Fang~\etal~\cite{fang2015captions} learn a visual attributes detector based on multi-instance learning (MIL) first and then learn a statistical language model for caption generation.
Likewise, Wu~\etal~\cite{wu2016value} train several visual attribute classifiers and take the outputs of those classifiers as inputs for the LSTM network to predict words.

In general, current recurrent neural network based approaches have shown their powerful capability on modeling word sequences~\cite{vinyals2015show,karpathy2015deep}.
However, the history-summarizing hidden states of RNNs are updated at each time, which render the long-term memory rather difficult~\cite{le2015simple, oliva2017statistical}.
Besides, we argue that current recurrent networks like LSTM are not efficient on modeling the hierarchical structure in word sequences.
All of these prompt us to explore a new language model to extract better sentence representation.
\textcolor{\MARK}{Considering} ConvNets can be stacked to extract hierarchical features over long-range contexts and have received a lot of attention in many tasks~\cite{gu2015recent}, in this paper, we design a language CNN to model words with long-term dependencies through multilayer ConvNets and to model the hierarchical representation through the bottom-up and convolutional architecture.

\begin{figure*}[ht]                    
	\centering                        
	\includegraphics[width=0.9\textwidth]{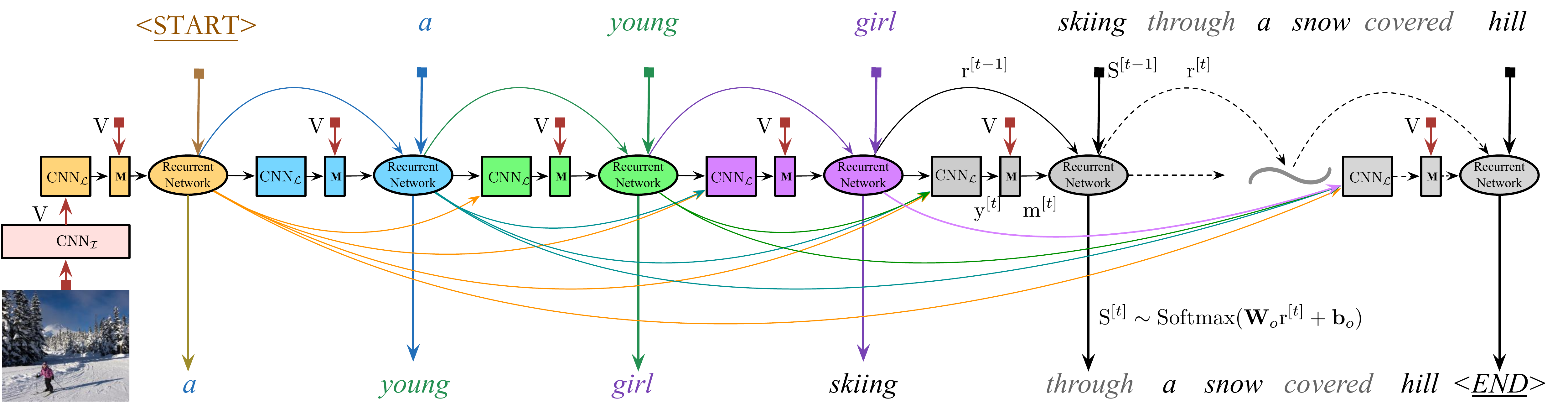}   
	\caption{An overview of our framework. The input of our model is a query image. Our model estimates the probability distribution of the next word given previous words and image. It consists of four parts: a $\text{CNN}_{\mathcal{I}}$ for image feature extraction, a deep $\text{CNN}_{\mathcal{L}}$ for language modeling, a multimodal layer ($\text{M}$) that connects the $\text{CNN}_{\mathcal{I}}$ and $\text{CNN}_{\mathcal{L}}$, and a Recurrent Network~(\eg, $\text{RNN}$, $\text{LSTM}$,~\etc) for word prediction. The weights are shared among all time frames.} 
	\label{fig:TOP} 
	\vspace{-5mm}
\end{figure*}
\section{Model Architecture}
\subsection{Overall Framework}
We study the effect of language CNN by combining it with Recurrent Networks.
Figure~\ref{fig:TOP} shows a recursive framework.
It consists of one deep CNN for image encoding, one CNN for sentence modeling, and a recurrent network for sequence prediction.
In order to distinguish these two CNN networks, we name the first CNN for image feature extraction as $\text{CNN}_\mathcal{I}$, and the second CNN for language modeling as $\text{CNN}_{\mathcal{L}}$.

Given an image $\mathbf{I}$, we take the widely-used CNN architecture VGGNet (16-layer)~\cite{simonyan2014very} pre-trained on ImageNet~\cite{krizhevsky2012imagenet} to extract the image features $\mathrm{V}\in \mathbb{R}^{K}$.
The $\text{CNN}_{\mathcal{L}}$ is designed to represent words and their hierarchical structure in word sequences.
It takes a sequence of $t$ generated words (each word is encoded as a one-hot representation) as inputs and generates a bottom-up representation of these words.
The outputs of $\text{CNN}_{\mathcal{I}}$ and $\text{CNN}_{\mathcal{L}}$ will be fed into a multimodal fusion layer, and use the recurrent network $f_{\text{recurrent}}(\cdot)$ to predict the next word.
The following equations show the main working flow of our model:
\begin{align}
\label{eq:ICNN}
\mathrm{V} &= \,\,\quad \text{CNN}_{\mathcal{I}}(\mathbf{I})\\
\label{eq:L_CNN}
\mathrm{y}^{[t]} &= \,\,\quad\text{CNN}_{\mathcal{L}}(\mathrm{S}^{[0]},\mathrm{S}^{[1]},\cdots,\mathrm{S}^{[t-1]})\\
\label{eq:CNN_M}
\mathrm{m}^{[t]}&= \,\,\quad\ f_{\text{multimodal}}(\mathrm{y}^{[t]},\mathrm{V})\\
\label{eq:CNN_MRHN}
\mathrm{r}^{[t]}&= \,\,\quad\ f_{\text{recurrent}}(\mathrm{r}^{[t-1]}, \mathrm{x}^{[t-1]}, \mathrm{m}^{[t]})\\
\label{eq:CNN_PREDICTION}
\mathrm{S}^{[t]}&\sim \,\,\quad \arg \max_{\mathcal{S}} \text{Softmax}(\mathbf{W}_o\mathrm{r}^{[t]}+\mathbf{b}_o)
\end{align}
where $t\in [0,N-1]$ is the time step, $\mathrm{y}^{[t]}$ is the output vector of $\text{CNN}_\mathcal{L}$, $\mathrm{r}^{[t]}$ is the activation output of recurrent network, $\mathrm{S}^{[t]}$ is the $t$-th word drawn from the dictionary $\mathcal{S}$ according to the maximum Softmax probability controlled by $\mathrm{r}^{[t]}$, $\mathbf{W}_o$ and $\mathbf{b}_o$ are weights and biases used for calculating the distribution over words.
Equation~\ref{eq:L_CNN},~\ref{eq:CNN_M},~\ref{eq:CNN_MRHN} and~\ref{eq:CNN_PREDICTION} are recursively applied, the design of each function is discussed below.

\subsection{$\text{CNN}_\mathcal{L}$ Layer}\label{CNN_L}
Models based on RNNs have dominated recent sequence modeling tasks~\cite{kuen2016recurrent, liu2016spatio,liu2017global, sutskever2014sequence}, and most of the recent image captioning models are based on LSTM networks~\cite{donahue2015long, karpathy2015deep, lu2016knowing}.
However, LSTM networks cannot explicitly model the hierarchical representation of words.
Even with multi-layer LSTM networks, such hierarchical structure is still hard to be captured due to the more complex model and higher risk of over-fitting.

\begin{figure}[ht]
	\begin{center}
		\centerline{\includegraphics[width=\columnwidth]{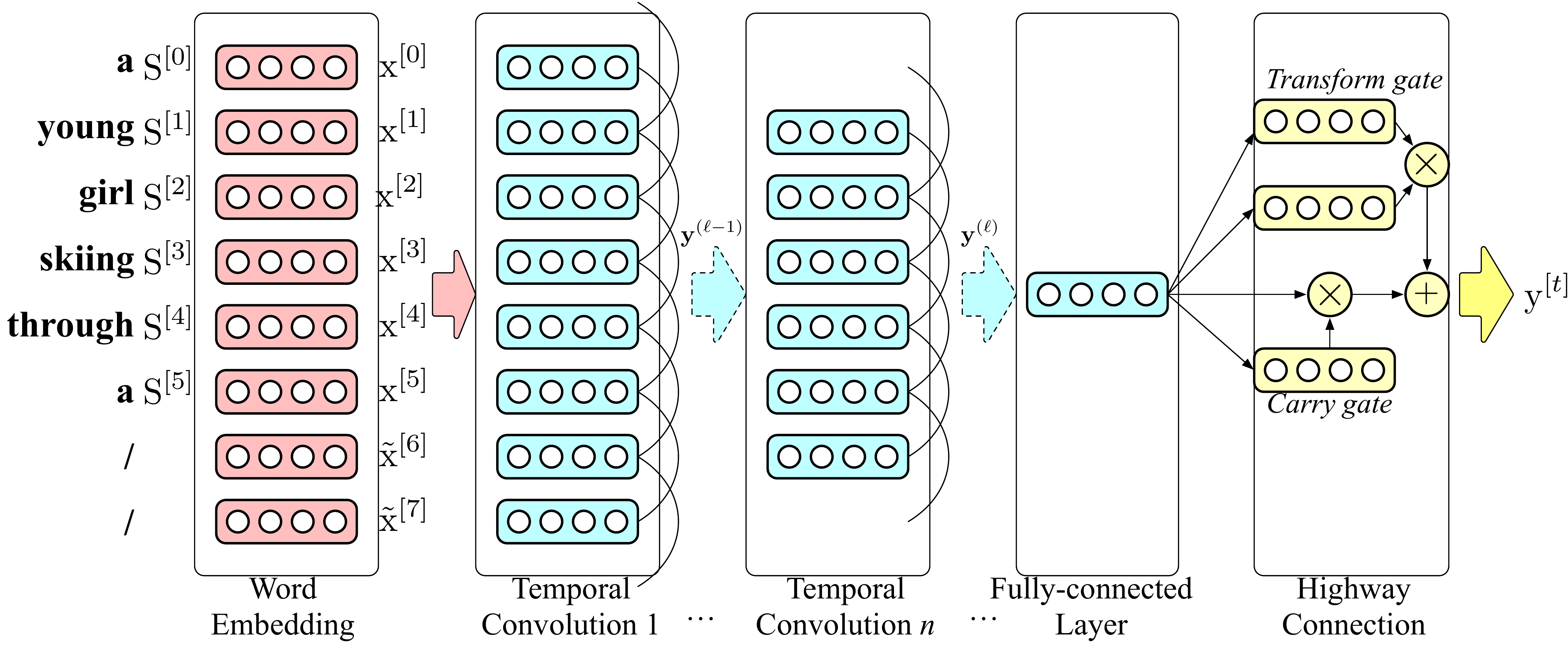}}
		\caption{The architecture of language CNN for sentence modeling. Here ``/" stands for a zero padding. The $\text{CNN}_\mathcal{L}$ builds a hierarchical representation of history words which contains the useful information for next word prediction.}
		\label{fig:LCNN_core}
	\end{center}
	\vspace{-5mm}
\end{figure}

Inspired by the recent success of CNNs in computer vision~\cite{gu2015recent, hudeep}, we adopt a language CNN with a hierarchical structure to capture the long-range dependencies between the input words, called $\text{CNN}_\mathcal{L}$.
The first layer of $\text{CNN}_\mathcal{L}$ is a word embedding layer.
It embeds the one-hot word encoding from the dictionary into word representation through a lookup table.
Suppose we have $t$ input words $\mathbf{S}=\{\mathrm{S}^{[0]},\mathrm{S}^{[1]},\cdots,\mathrm{S}^{[t-1]}\}$, and $\mathrm{S}^{[i]}$ is the one-of-$V$ (one-hot) encoding, with $V$ as the size of the vocabulary.
We first map each word $\mathrm{S}^{[t]}$ in the sentence into a $K$-dimensional vector $\mathrm{x}^{[t]}=\mathbf{W}_e \mathrm{S}^{[t]}$, where $\mathbf{W}_e\in \mathbb{R}^{K\times V}$ is a word embedding matrix (to be learned).
Next, those embeddings are concatenated to produce a matrix as follows:
\begin{equation}\label{eq:L_CNN_SENTENCE}
\mathbf{x} =\left[\mathrm{x}^{[0]},\mathrm{x}^{[1]},\cdots,\mathrm{x}^{[t-1]}\right]^T, \mathbf{x}\in \mathbb{R}^{t \times K}
\end{equation}
The concatenated matrix $\mathbf{x}$ is fed to the convolutional layer.
Just like the normal CNN, $\text{CNN}_\mathcal{L}$ has a fixed architecture with predefined maximum number of input words~(denoted as $L_{\mathcal{L}}$).
Unlike the toy example in Figure~\ref{fig:LCNN_core}, in practice we use a larger and deeper $\text{CNN}_\mathcal{L}$ with $L_\mathcal{L}=16$. 

We use the temporal convolution~\cite{kiros2014multimodal} to model the sentence.
Given an input feature map $\mathbf{y}^{(\ell-1)}\in \mathbb{R}^{M_{\ell-1}\times K}$ of Layer-$\ell-1$, the output feature map $\mathbf{y}^{(\ell)}\in \mathbb{R}^{M_{\ell}\times K}$ of the temporal convolution layer-$\ell$ will be:
\begin{equation}\label{eq:TCONV}
{\mathrm{y}_{i}^{(\ell)}}(\mathbf{x})=\sigma (\mathbf{w}^{(l)}_{L}{\mathbf{y}}_i^{(\ell-1)}+\mathrm{b}_L^{(\ell)})
\end{equation}
here ${\mathrm{y}_{i}^{(\ell)}}(\mathbf{x})$ gives the output of feature map for location $i$ in Layer-$\ell$, $\mathbf{w}_{L}^{(l)}$ denotes the parameters on Layer-$\ell$, $\sigma(\cdot)$ is the activation function,~\eg, Sigmoid, or ReLU.
The input feature map ${\mathbf{y}}_i^{(l-1)}$ is the segment of Layer-$\ell-1$ for the convolution at location $i$, while $\mathbf{y}^{(0)}$ is the concatenation of $t$ word embeddings from the sequence input $\mathrm{S}^{[0:t-1]}$.
The definition of $\mathbf{y}^{(0)}$ is as follows:
\begin{equation}\label{input_cnn_0}
\mathbf{y}^{(0)}\overset{\text{def}}{=}\begin{cases}
\left[\mathrm{x}^{[t-L_{\mathcal{L}}]},\cdots,\mathrm{x}^{[t-1]}\right]^T, &\text{if } t\ge L_{\mathcal{L}}\\
\left[\mathrm{x}^{[0]},\cdots,\mathrm{x}^{[t-1]},\tilde{\mathrm{x}}^{[t]},\cdots,\tilde{\mathrm{x}}^{[L_{\mathcal{L}}-1]}\right]^T &\text{otherwise}
\end{cases}
\end{equation}
Specially, when $t\ge L_{\mathcal{L}}$, the input sentence will be truncated, we only use $L_{\mathcal{L}}$ words before the current time step $t$.
When $t<L_{\mathcal{L}}$, the input sentence will be padded with $\tilde{\mathrm{x}}^{[:]}$.
Note that if $t=0$, $\tilde{\mathrm{x}}^{[:]}$ are the image features $\mathrm{V}$, otherwise $\tilde{\mathrm{x}}^{[:]}$ are the zero vectors that have the same dimension as $\mathrm{x}^{[:]}$.

Previous CNNs, including those adopted for NLP tasks~\cite{hu2014convolutional,kalchbrenner2014convolutional}, take the classic \textit{convolution-pooling} strategy, which uses \textit{max-pooling} to pick the highest response feature across time.
This strategy works well for tasks like text classification~\cite{kalchbrenner2014convolutional} and matching~\cite{hu2014convolutional}, but is undesirable for modeling the composition functionality, because it ignores the temporal information in sequence.
In our network, we discard the pooling operations.
We consider words as the smallest linguistic unit and apply a straightforward stack of convolution layers on top of each other.
In practice, we find that deeper $\text{CNN}_\mathcal{L}$ works better than shallow $\text{CNN}_\mathcal{L}$, which is consistent with the tradition of CNNs in computer vision~\cite{gu2015recent}, where using very deep CNNs is key to having better feature representation.

The output features of the final convolution layer are fed into a fully connected layer that projects the extracted words features into a low-dimensional representation.
Next, the projected features will be fed to a highway connection~\cite{srivastava2015training} which controls flows of information in the layer and improves the gradient flow.
The final output of the highway connection is a $K$-dimensional vector $\mathrm{y}^{[t]}$.

\subsection{Multimodal Fusion Layer}
Next, we add a multimodal fusion layer after $\text{CNN}_\mathcal{L}$, which fuses words representation and image features.
This layer has two inputs: the bottom-up words representation $\mathrm{y}^{[t]}$ extracted from $\text{CNN}_{\mathcal{L}}$ and the image representation $\mathrm{V}$ extracted from $\text{CNN}_{\mathcal{I}}$.
We map these two inputs to the same multimodal feature space and combine them together to obtain the activation of multimodal features:
\begin{align}
\mathrm{m}^{[t]}&=f_{\text{multimodal}}(\mathrm{y}^{[t]},\mathrm{V}) \label{eq:multimodal}\\
&=\sigma \left( f_{\mathrm{y}}(\mathrm{y}^{[t]};\mathbf{W}_{\mathrm{Y}},\mathbf{b}_{\mathrm{Y}})  + g_\mathrm{v}( \mathrm{V};\mathbf{W}_{\mathrm{V}},\mathbf{b}_{\mathrm{V}})\right) 
\end{align}
where ``+" denotes element-wise addition, $f_{\mathrm{y}}(\cdot)$ and $g_{\mathrm{v}}(\cdot)$ are linear mapping functions, $\mathrm{m}^{[t]}$ is the multimodal layer output feature vector.
$\sigma(\cdot)$ is the activation function, here we use the scaled $\tanh$ function~\cite{lecun2012efficient} which leads to a faster training process than the basic $\tanh$ function.

\subsection{Recurrent Networks}\label{sec:rr}
Our $\text{CNN}_\mathcal{L}$ may miss the important temporal information because it extracts the holistic features from the whole sequence of words.
To overcome this limitation, we combine it with recurrent networks.
In our model, the transition equations of the recurrent network can be formulated as:
\begin{align}
\mathrm{r}^{[t]}&= f_{\text{recurrent}}(\mathrm{r}^{[t-1]}, \mathrm{x}^{[t-1]}, \mathrm{m}^{[t]})\label{eq:recurrent}\\
\mathrm{S}^{[t]}&\sim \arg \max_{\mathcal{S}} \text{Softmax}(\mathbf{W}_o\mathrm{r}^{[t]}+\mathbf{b}_o)
\end{align}
where $\mathrm{r}^{[t]}$ denotes the recurrent state, $\mathrm{x}^{[t-1]}=\mathbf{W}_e \mathrm{S}^{[t-1]}$ is the previous word embedding, $\mathrm{m}^{[t]}$ is the multimodal fusion output, and $f_{\text{recurrent}}(\cdot)$ is the transition function of recurrent network.
$\text{Softmax}(\mathrm{r}^{[t]})$ is the probability of word $\mathrm{S}^{[t]}$ given by the Softmax layer, and $\mathrm{S}^{[t]}$ is the $t$-th decoded word.
In our study, we combine our language CNN with four types of recurrent networks: Simple RNN, LSTM network, GRU~\cite{cho2014learning}, and Recurrent Highway Network (RHN)~\cite{zilly2016recurrent}.

Traditionally, the simple RNN updates the recurrent state $\mathrm{r}^{[t]}$ of Equation~\ref{eq:recurrent} as follows:
\begin{equation}
\mathrm{r}^{[t]}=\tanh(\mathbf{W}_r\mathrm{r}^{[t-1]}+\mathbf{W}_z\mathrm{z}^{[t]}+\mathbf{b})\label{eq:simpleRNN}
\end{equation}
where $\mathrm{z}^{[t]}$ is the input.
However, this type of simple RNN is hard to deal with long-term dependencies~\cite{bengio1994learning}.
As the vanishing gradient will make gradients in directions that short-term dependencies are large, while the gradients in directions that correspond to long-term dependencies are small.

LSTM network extents the simple RNN with the gating mechanism (\textit{input} gate, \textit{forget} gate, and \textit{output} gate) to control information flow and a \textit{memory cell} to store the history information, thus it can better model the long-term dependencies than simple RNN.

GRU is an architecture similar to the LSTM, but it has a simplified structure.
GRU does not has a separate \textit{memory cell} and exposes its hidden state $\mathrm{r}^{[t]}$ without any control.
Thus, it is computationally more efficient and outperforms the LSTM network on many tasks due to its simple structure.


Besides, we also consider a fourth type of recurrent network: RHN, which introduces the highway connection to simple RNN.
RHN has directly gated connections between previous state $\mathrm{r}^{[t-1]}$ and current input $\mathrm{z}^{[t]}$ to modulate the flow of information.
The transition equations of RHN can be formulated as follows:
\begin{eqnarray}
\left(
\begin{array}{ccc}
\mathrm{t}^{[t]} \\ \mathrm{c}^{[t]}\\ \mathrm{h}^{[t]} \\ 
\end{array}
\right)
&=&
\left(
\begin{array}{ccc}
\sigma \\ \sigma \\\tanh \\ 
\end{array}
\right)
\left(
\mathbf{M}
\left(
\begin{array}{ccc}
\mathrm{r}^{[t-1]} \\ \mathrm{z}^{[t]} \\
\end{array}
\right)
\right)\\
\mathrm{r}^{[t]} &=& \mathrm{h}^{[t]} \odot \mathrm{t}^{[t]}  + \mathrm{c}^{[t]} \odot  \mathrm{r}^{[t-1]}
\label{eq:rhn_ht}
\end{eqnarray}
where $\mathrm{c}^{[t]}$ is the \textit{carry} gate, $\mathrm{t}^{[t]}$ is the \textit{transform} gate, $\mathrm{h}^{[t]}$ denotes the modulated input, $\mathbf{M}:\mathbb{R}^{2K+d}\rightarrow\mathbb{R}^{3d}$ is an affine transformation.
$\mathrm{z}^{[t]}\in \mathbb{R}^{2K}$ denotes the concatenation of two vectors: $\mathrm{m}\ts{t}$ and $\mathrm{x}\ts{t-1}$.
According to Equation~\ref{eq:CNN_M} and~\ref{eq:L_CNN}, $\mathrm{z}^{[t]}$ can be expressed as follows:
\begin{align}
\mathrm{z}^{[t]}=[f_{\text{multimodal}}({\text{CNN}_{\mathcal{L}}}(\mathrm{x}^{[0,\cdots,t-1]}),\mathrm{V});\mathrm{x}^{[t-1]}]
\label{eq:zt}
\end{align}

Like GRU, RHN does not have \textit{output} gate to control the exposure of the recurrent state $\mathrm{r}^{[t]}$, but exposes the whole state each time.
The RHN, however, does not have \textit{reset} gate to drop information that is irrelevant in the future.
As our $\text{CNN}_{\mathcal{L}}$ can extract the relevant information from the sequence of history words at each time step, to some extent, the $\text{CNN}_{\mathcal{L}}$ allows the model to add information that is useful in making a prediction.

\subsection{Training}
During training, given the ground truth words $\mathbf{S}$ and corresponding image $\mathbf{I}$, the loss function for a single training instance $(\mathbf{S},\mathbf{I})$ is defined as a sum of the negative log likelihood of the words.
The loss can be written as:
\begin{equation}\label{eq:loss_function}
\mathcal{L}(\mathbf{S},\mathbf{I}) =-\sum_{t=0}^{N-1} \log P(\mathrm{S}^{[t]}|\mathrm{S}^{[0]},\cdots,\mathrm{S}^{[t-1]},\mathbf{I})
\end{equation}
where $N$ is the sequence length, and $\mathrm{S}^{[t]}$ denotes a word in the sentence $\mathbf{S}$. 

The training objective is to minimize the cost function, which is equivalent to maximizing the probability of the ground truth context words given the image by using: $\arg\max_{\theta}\sum_{t=0}^{N-1} \log P(\mathrm{S}^{[t]}|\mathrm{S}^{[0:t-1]},\mathbf{I})$, where $\theta$ are the parameters of our model, and $P(\mathrm{S}^{[t]}|\mathrm{S}^{[0:t-1]},\mathbf{I})$ corresponds to the activation of Softmax layer.

\subsection{Implementation Details}
In the following experiments, we use the 16-layer VGGNet~\cite{simonyan2014very} model to compute CNN features and map the last fully-connected layer's output features to an embedding space via a linear transformation.

As for preprocessing of captions, we transform all letters in the captions to lowercase and remove all the non-alphabetic characters.
Words occur less than five times are replaced with an unknown token $<$UNK$>$.
We truncate all the captions longer than 16 tokens and set the maximum number of input words for $\text{CNN}_{\mathcal{L}}$ to be 16. 

\subsubsection{Training Details}
In the training process, each image $\mathbf{I}$ has five corresponding annotations.
We first extract the image features $\mathrm{V}$ with $\text{CNN}_{\mathcal{I}}$.
The image features $\mathrm{V}$ are used in each time step.
We map each word representation $\mathrm{S}^{[t]}$ with: $\mathrm{x}^{[t]}=W_e \mathrm{S}^{[t]} , t\in [0,N-1]$.
After that, our network is trained to predict the words after it has seen the image and preceding words.
Please note that we denote by $\mathrm{S}^{[0]}$ a special $<$START$>$ token and by $\mathrm{S}^{[N-1]}$ a special $<$END$>$ token which designate the start and end of the sentence.

For Flickr30K~\cite{plummer2015flickr30k} and MS COCO~\cite{lin2014microsoft} we set the dimensionality of the image features and word embeddings as 512.
All the models are trained with Adam~\cite{kingma2014adam}, which is a stochastic gradient descent method that computes adaptive learning rate for each parameter.
The learning rate is initialized with 2e-4 for Flickr30K and 4e-4 for MS COCO, and the restart technique mentioned in~\cite{loshchilov2016sgdr} is adopted to improve the convergence of training.
Dropout and early stopping are used to avoid overfitting.
All weights are randomly initialized except for the CNN weights.
More specifically, we fine-tune the VGGNet when the validation loss stops decreasing.
The termination of training is determined by evaluating the CIDEr~\cite{vedantam2015cider} score for the validation split after each training epoch.

\subsubsection{Testing}
During testing, the previous output $\mathrm{S}^{[t-1]}$ is used as input in lieu of $\mathrm{S}^{[t]}$.
The sentence generation process is straightforward.
Our model starts from the $<$START$>$ token and calculates the probability distribution of the next word : $P(\mathrm{S}^{[t]}|\mathrm{S}^{[0:t-1]},\mathbf{I})$.
Here we use Beam Search technology proposed in~\cite{jia2015guiding}, which is a fast and efficient decoding method for recurrent network models. We set a fixed beam search size ($k$=2) for all models (with RNNs) in our tests. 

\section{Experiments}
\subsection{Datasets and Evaluation Metrics}
We perform experiments on two popular datasets that are used for image caption generation: MS COCO and Flickr30k.
These two datasets contain 123,000 and 31,000 images respectively, and each image has five reference captions.
For MS COCO, we reserve 5,000 images for validation and 5,000 images for testing.
For Flickr30k, we use 29,000 images for training, 1,000 images for validation, and 1,000 images for testing. 

We choose four metrics for evaluating the quality of the generated sentences:
\textbf{BLEU-$n$}~\cite{papineni2002bleu} is a precision-based metric.
It measures how many words are shared by the generated captions and ground truth captions.
\textbf{METEOR}~\cite{denkowski2014meteor} is based on the explicit word to word matches between generated captions and ground-truth captions.
\textbf{CIDEr}~\cite{vedantam2015cider} is a metric developed specifically for evaluating image captions.
It measures consensus in image caption by performing a Term Frequency-Inverse Document Frequency weighting for each $n$-gram.
\textbf{SPICE}~\cite{anderson2016spice} is a more recent metric which has been shown to correlate better with the human judgment of semantic quality than previous metrics.

\subsection{Models}\label{BASE_LINE}
To gain insight into the effectiveness of $\text{CNN}_{\mathcal{L}}$, we compare $\text{CNN}_{\mathcal{L}}$-based models with methods using the recurrent network only.
For a fair comparison, the output dimensions of all gates are fixed to 512.

\textbf{Recurrent Network-based Models.}
We implement Recurrent Network-based Models based on the framework proposed by Vinyals~\etal~\cite{vinyals2015show}, it takes an image as input and predicts words with one-layer Recurrent Network.
Here we use the publicly available implementation Neuraltalk2~\footnote{\url{https://github.com/karpathy/neuraltalk2}}.
We evaluate four baseline models: \textbf{Simple RNN}, \textbf{RHN}, \textbf{LSTM}, and \textbf{GRU}.

\textbf{$\text{CNN}_{\mathcal{L}}$-based Models.}
As can be seen in Figure~\ref{fig:TOP}.
The $\text{CNN}_{\mathcal{L}}$-based models employ a $\text{CNN}_{\mathcal{L}}$ to obtain the bottom-up representation from the sequence of words and cooperate with the Recurrent Network to predict the next word.
Image features and words representation learned from $\text{CNN}_{\mathcal{I}}$ and $\text{CNN}_{\mathcal{L}}$ respectively are fused with the multimodal function.
We implement four $\text{CNN}_{\mathcal{L}}$-based models: \textbf{$\text{CNN}_{\mathcal{L}}$+Simple RNN}, \textbf{$\text{CNN}_{\mathcal{L}}$+RHN}, \textbf{$\text{CNN}_{\mathcal{L}}$+LSTM}, and \textbf{$\text{CNN}_{\mathcal{L}}$+GRU}.

\subsection{Quantitative Results}
We first evaluate the importance of language CNN for image captioning, then evaluate the effects of $\text{CNN}_{\mathcal{L}}$ on two datasets (Flickr30K and MS COCO), and \textcolor{\MARK}{also compare with the state-of-the-art methods.}

\subsubsection{Analysis of $\text{CNN}_{\mathcal{L}}$}
\textcolor{\MARK}{It is known that} $\text{CNN}_{\mathcal{L}}$-based models have \textcolor{\MARK}{larger} capacity than RNNs.
To verify that the improved performance is from the developed $\text{CNN}_{\mathcal{L}}$ rather than due to more layers/parameters, we set the hidden and output sizes of RNNs to 512 and 9568 (vocabulary size), and list the parameters of each model as well as their results in Table~\ref{tab:basic_comparsion_params}.

\begin{table}[ht]
	\centering
	\setlength{\tabcolsep}{2pt}
	\scalebox{0.95}{
		{\small
			\begin{tabular}{  l |r | c | c |  l |r | c  | c }
				\hline
				Approach&Params & B@4 & C &  Approach & Params & B@4 & C\\
				\hline
				{Simple RNN} & 5.4M &27.0 &	87.0  & {LSTM} & 7.0M	& 29.2 &92.6 \\
				{$\text{CNN}_{\mathcal{L}}$} & 6.3M & 18.4 & 56.8 & {$\text{LSTM}_{\text{2}}$} & 9.1M & \textbf{29.7}  & 93.2 \\
				{$\text{CNN}_{\mathcal{L}}$+RNN} &\textbf{11.7M} &29.5 &  \textbf{95.2} &{$\text{LSTM}_{\text{3}}$}  & 11.2M& 29.3  &	92.9 \\
				\hline
			\end{tabular}
		}
	}
	\caption{Results on MS COCO, where B@n are short for BLEU-n, C is short for CIDEr. All values are reported as percentage (Bold numbers are the best results). $\text{CNN}_{\mathcal{L}}$ contains five temporal convolutional layers, the kernel size of the first two convolutional layers is 5, and the rest kernel size of convolutional layers is 3.}
	\label{tab:basic_comparsion_params}
	\vspace{-5mm}
\end{table}

As seen in Table~\ref{tab:basic_comparsion_params}, the parameter size of the 3-layer LSTM ($\text{LSTM}_{3}$) is close to that of the $\text{CNN}_{\mathcal{L}}$+RNN. \textcolor{\MARK}{Adding} the $2^{\text{nd}}$ LSTM layer ($\text{LSTM}_{2}$) improves the performance of LSTM, but it is still lower than $\text{CNN}_{\mathcal{L}}$+RNN.
Meanwhile, $\text{LSTM}_{3}$ does not show improvements as the model experiences overfitting.
This issue is even worse on Flickr30K which has relatively small number of training data.
Note that $\text{CNN}_{\mathcal{L}}$ (without RNNs) achieves lower performance than $\text{CNN}_{\mathcal{L}}$+RNN.
We find that those predicted captions of $\text{CNN}_{\mathcal{L}}$ (without RNNs) only are short, but contain primary attributes,~\eg, $\text{CNN}_{\mathcal{L}}$ model generates: ``\textit{a person on a wave}", while $\text{CNN}_{\mathcal{L}}$+RNN provides: ``\textit{a young man surfing a wave}".
This finding shows that the temporal recurrence of RNNs is \textcolor{\MARK}{still} crucial for modeling the short-term contextual information across words in the sentence.

\begin{table}[t!]
	\centering
	\setlength{\tabcolsep}{2pt}
	\scalebox{0.95}{
		{\small
			\begin{tabular}{l | c | c | l | c | c}
				\hline
				Approach& B@4 & C &  Approach& B@4 & C\\
				\hline
				{$\text{Avg}_{\text{history}}$+RHN}  & 30.1  & 95.8 & {$\text{CNN}_{\mathcal{L}_{\text{2 words}}}$+RHN} & 29.2 & 93.8\\
				{$\text{CNN}_{\mathcal{L}_{\text{16 words}}^{\ast}}$+RHN} & 28.9 & 91.9 & {$\text{CNN}_{\mathcal{L}_{\text{4 words}}}$+RHN} &29.5 & 95.8\\
				{$\text{CNN}_{\mathcal{L}}$+RHN} & \textbf{30.6} & \textbf{98.9} & {$\text{CNN}_{\mathcal{L}_{\text{8 words}}}$+RHN} & 30.0 & 95.9\\
				\hline
			\end{tabular}
		}
	}
	\caption{Results of different history information encoding approaches on MS COCO. $\text{CNN}_{\mathcal{L}_{N \text{words}}}$ takes $N$ previous words as inputs, \textcolor{\MARK}{where} we set $N$ to 2, 4, and 8. $\text{Avg}_{\text{history}}$ computes an average over history word embeddings.  $\text{CNN}_{\mathcal{L}_{\text{16 words}}^{\ast}}$ replaces the  $2^{\text{nd}}$ and  $4^{\text{th}}$ convolutional layers in $\text{CNN}_{\mathcal{L}}$ with the \textit{max-pooling} layer.}
	\label{tab:basic_comparsion_cnn_l}
	\vspace{-5mm}
\end{table}

We further compare language CNNs with different input words and with \textit{max-pooling} operations, \textcolor{\MARK}{where those language CNNs are combined with RHN instead of RNN}.
Table~\ref{tab:basic_comparsion_cnn_l} shows that larger context windows achieve better performance.
\textcolor{\MARK}{This is likely because} $\text{CNN}_{\mathcal{L}}$ with larger window size can better utilize contextual information and learn better word embedding representation.
In addition, the performance of $\text{CNN}_{\mathcal{L}_{\text{16 words}}^{\ast}}$+RHN is inferior to $\text{CNN}_{\mathcal{L}}$+RHN, which experimentally supports our opinion that \textit{max-pooling} operations lose information about the local order of words. 

\subsubsection{Results Using $\text{CNN}_{\mathcal{L}}$ on MS COCO}\label{RESULT_CNNl}
Table~\ref{tab:coco_compare} shows the generation performance on MS COCO.
By combine $\text{CNN}_{\mathcal{L}}$, our methods clearly outperforms the recurrent network counterpart in all metrics.
\begin{table}[ht]
	\centering
	\setlength{\tabcolsep}{2pt}
	\scalebox{0.95}{
		{\small
			\begin{tabular}{l | c c c c c c c}
				\hline
				Approach& B@1 & B@2 & B@3 & B@4 & M & C & S\\
				\hline
				{Simple RNN} & 70.1	& 52.1	& 37.6	& 27.0 & 23.2 &	87.0 & 16.0\\
				{$\text{CNN}_{\mathcal{L}}$+RNN} & 72.2 & 55.0 & 40.7 & 29.5 & 24.5 & 95.2 & 17.6 \\
				\hline
				{RHN} & 70.5	& 52.7	& 37.8	& 27.0 & 24.0 &	90.6 & 17.2\\
				{$\text{CNN}_{\mathcal{L}}$+RHN} & 72.3 & 55.3 & \textbf{41.3} & \textbf{30.6} & \textbf{25.2} & 98.9 & \textbf{18.3}\\
				\hline
				{LSTM} & 70.8	& 53.6	& 39.5	& 29.2 & 24.5 &	92.6 & 17.1\\
				{$\text{CNN}_{\mathcal{L}}$+LSTM} & 72.1 & 54.6 & 40.9 & 30.4 & 25.1 & \textbf{99.1} & 18.0\\
				\hline
				{GRU} & 71.6	& 54.1	& 39.7	& 28.9 & 24.3 &	93.3 & 17.2\\
				{$\text{CNN}_{\mathcal{L}}$+GRU} & \textbf{72.6} & \textbf{55.4} & 41.1 & 30.3 & 24.6 & 96.1 & 17.6\\
				\hline
			\end{tabular}
		}
	}
	\caption{Performance comparison on MS COCO, where M is short for METEOR, and S is short for SPICE.}
	\label{tab:coco_compare}
	\vspace{-2mm}
\end{table}

\begin{table}[ht]
	\centering
	\setlength{\tabcolsep}{2pt}
	\scalebox{0.95}{
		{\small
			\begin{tabular}{l | c c  c c c c c }
				\hline
				Approach& B@1 & B@2 & B@3 & B@4 & M & C & S\\
				\hline
				Simple RNN & 60.5	& 41.3	& 28.0	& 19.1 & 17.1 &	32.5 & 10.5\\
				$\text{CNN}_{\mathcal{L}}$+RNN & {71.3} & {53.8} & {39.6} & 28.7 & \textbf{22.6} & \textbf{65.4} & \textbf{15.6} \\
				\hline
				RHN & 62.1 & 43.1 & 29.4 & 20.0 & 17.7 & 38.4 & 11.4\\
				$\text{CNN}_{\mathcal{L}}$+RHN 				& \textbf{73.8} & \textbf{56.3} & \textbf{41.9} & \textbf{30.7} & 21.6 & 61.8 & 15.0 \\
				\hline
				LSTM & 60.9 & 41.8 & 28.3 & 19.3 & 17.6 & 35.0 & 11.1 \\
				$\text{CNN}_{\mathcal{L}}$+LSTM & 64.5 & 45.8 & 32.2 & 22.4 & 19.0  & 45.0 & 12.5 \\
				\hline 
				GRU & 61.4 & 42.5 & 29.1 & 20.0 & 18.1 & 39.5 & 11.4\\
				$\text{CNN}_{\mathcal{L}}$+GRU 				& 71.4 & 54.0 & 39.5 & 28.2 & 21.1 & 57.9 & 14.5\\
				\hline
			\end{tabular}
		}
	}
	\caption{Performance comparison on Flickr30k.}
	\label{tab:Flickr30k_compare}
	\vspace{-2mm}
\end{table}
\begin{table*}[h!]
	\centering
	\setlength{\tabcolsep}{2pt}
	\scalebox{0.95}{
		{\small
			\begin{tabular}{l | c c c c c c | c c c c c c} 
				\hline
				& \multicolumn{5}{c}{\textit{Flickr30k}} & & \multicolumn{6}{c}{\textit{MS COCO}} \\
				\cline{2-13}
				Approach & BLEU-1 &BLEU-2 & BLEU-3 & BLEU-4 & METEOR & & BLEU-1 &BLEU-2 &BLEU-3 &BLEU-4 & METEOR & CIDEr\\
				\hline 
				\textit{BRNN~\cite{karpathy2015deep}} & 57.3 & 36.9 & 24.0 & 15.7 & \textemdash	& & 62.5 & 45.0	& 32.1	& 23.0 & 19.5 &	66.0 \\
				\textit{Google NIC~\cite{vinyals2015show}}      & \textemdash & \textemdash & \textemdash & \textemdash & \textemdash 	& & \textemdash	& \textemdash	& \textemdash	& 27.7 & 23.7 &	85.5 \\  
				\textit{LRCN~\cite{donahue2015long}}   			& 58.8 & 39.1 & 25.1 & 16.5 & \textemdash  	& & 66.9 & 48.9 & 34.9 & 24.9 & \textemdash & \textemdash \\
				\textit{MSR~\cite{fang2015captions}}  & \textemdash & \textemdash & \textemdash & \textemdash & \textemdash && \textemdash	& \textemdash	& \textemdash 	& 25.7 & 23.6 &	\textemdash \\
				\textit{m-RNN~\cite{mao2014deep}}       		& 60.0 & 41.0 & 28.0 & 19.0 & \textemdash 	& & 67.0 & 49.0 & 35.0 & 25.0 & \textemdash &	\textemdash\\
				\textit{Hard-Attention~\cite{xu2015show}}       & 66.9 & 43.9 & 29.6 & 19.9 & 18.5 & & 70.7	& 49.2	& 34.4	& 24.3 & 23.9 &	\textemdash \\
				\textit{Soft-Attention~\cite{xu2015show}}       & 66.7 & 43.4 & 28.8 & 19.1 & 18.5 & & 71.8	& 50.4	& 35.7	& 25.0 & 23.0 &	\textemdash \\
				\textit{ATT-FCN~\cite{you2016image}} 			& 64.7 & 46.0 & 32.4 & 23.0 & 18.9 & & 70.9	& 53.7	& 40.2	& 30.4 & 24.3 &	\textemdash \\
				\textit{ERD+GoogLeNet~\cite{yang2016encode}}  & \textemdash & \textemdash & \textemdash & \textemdash & \textemdash && \textemdash	& \textemdash	& \textemdash 	& 29.8 & 24.0 &	88.6 \\
				\textit{emb-gLSTM~\cite{jia2015guiding}}		& 64.6 & 44.6 & 30.5 & 20.6 & 17.9 && 67.0	& 49.1	& 35.8	& 26.4 & 22.7 &	81.3 \\
				\textit{VAE~\cite{pu2016variational}} 			& 72.0 & 53.0 & 38.0 & 25.0 & \textemdash & & 72.0	& 52.0	& 37.0	& 28.0 & 24.0 &	90.0\\
				\hline
				& \multicolumn{11}{c}{\textit{State-of-the-art results using model assembling or extra information}}\\
				\hline
				\textit{Google NICv2~\cite{vinyals2016show}}    & \textemdash & \textemdash & \textemdash & \textemdash & \textemdash & & \textemdash & \textemdash & \textemdash & {32.1} & {25.7} &	99.8	\\
				\textit{Attributes-CNN+RNN~\cite{wu2016value}} 			& 73.0 & 55.0 & 40.0 & 28.0 & \textemdash & & 74.0	& 56.0	& 42.0	& 31.0 & 26.0 &	94.0\\
				\hline
				& \multicolumn{11}{c}{\textit{Our results}}\\
				\hline
				{$\text{CNN}_{\mathcal{L}}$+RNN }  & 71.3 & 53.8 & 39.6 & 28.7 & 22.6 && 72.2 & 55.0 & 40.7 & 29.5 & 24.5 & 95.2\\
				{$\text{CNN}_{\mathcal{L}}$+RHN}   & 73.8 & 56.3 & 41.9 & 30.7 & 21.6 && 72.3 & 55.3 & 41.3 & 30.6 & 25.2 & 98.9\\
				{$\text{CNN}_{\mathcal{L}}$+LSTM } & 64.5 & 45.8 & 32.2 & 22.4 & 19.0 && 72.1 & 54.6 & 40.9 & 30.4 & 25.1 & 99.1\\
				{$\text{CNN}_{\mathcal{L}}$+GRU }  & 71.4 & 54.0 & 39.5 & 28.2 & 21.1 && 72.6 & 55.4 & 41.1 & 30.3 & 24.6 & 96.1\\
				\hline
			\end{tabular}
		}
	}
	\caption{Performance in terms of BLEU-$n$, METEOR, and CIDEr compared with other state-of-the-art methods on the MS COCO and Flickr30k datasets. For those competing methods, we extract their performance from their latest version of papers.}
	\label{tab:state_of_the_art}
	\vspace{-5mm}
\end{table*}

Among these models, $\text{CNN}_{\mathcal{L}}$+RHN achieves the best performances in terms of B@(3,4), METEOR, and SPICE metrics, $\text{CNN}_{\mathcal{L}}$+LSTM achieves the best performance in CIDEr metric (99.1), and $\text{CNN}_{\mathcal{L}}$+GRU achieves the best performance in B@(1,2) metrics.
\textcolor{\MARK}{Although the absolute gains across different B@n metrics are similar, the percentage of the relative performance improvement is increasing from B@1 to B@4.}
It does show the advantage of our method in terms of better capturing long-term dependency.
Note that the $\text{CNN}_{\mathcal{L}}$+RNN model achieves better performance than simple RNN model and outperforms LSTM model.
As mentioned in Section~\ref{sec:rr}, LSTM networks model the word dependencies with multi-gates and the internal memory cell.
However, our $\text{CNN}_{\mathcal{L}}$+RNN without memory cell works better than LSTM model.
We think the reason is that our language CNN takes all history words as input and explicitly model the long-term dependencies in history words, this could be regarded as an external ``\textit{memory cell}".
Thus, the $\text{CNN}_{\mathcal{L}}$'s ability to model long-term dependencies can be taken as enhancement of simple RNNs, which can solve the difficulty of learning long-term dependencies.

\subsubsection{Results Using $\text{CNN}_{\mathcal{L}}$ on Flickr30K}\label{RESULT_CNN_30k}
We also evaluate the effectiveness of language CNN on the smaller dataset Flickr30K.
The results in Table~\ref{tab:Flickr30k_compare} clearly indicate the advantage of exploiting the language CNN to model the long-term dependencies in words for image captioning.
Among all models, $\text{CNN}_{\mathcal{L}}$+RHN achieves the best performances in B@(1,2,3,4) metrics, and $\text{CNN}_{\mathcal{L}}$+RNN achieves the best performances in METEOR, CIDEr, and SPICE metrics.

As for the low results (without $\text{CNN}_{\mathcal{L}}$) on Flickr30k, we think that it is due to lack of enough training data to avoid overfitting.
In contrast, our $\text{CNN}_{\mathcal{L}}$ can help learn better word embedding and \textcolor{\MARK}{better representation of history words} for word prediction, and it is much easier to be trained compared with LSTM due to its simplicity and efficiency.
Note that the performance of LSTM and $\text{CNN}_{\mathcal{L}}$+LSTM models are lower than RHN/GRU and $\text{CNN}_{\mathcal{L}}$+RHN/GRU.
This illustrates that the LSTM networks are easily overfitting on this smaller dataset. 

\begin{figure*}[ht]
	\begin{center}
		\centerline{\includegraphics[width=1.0\textwidth]{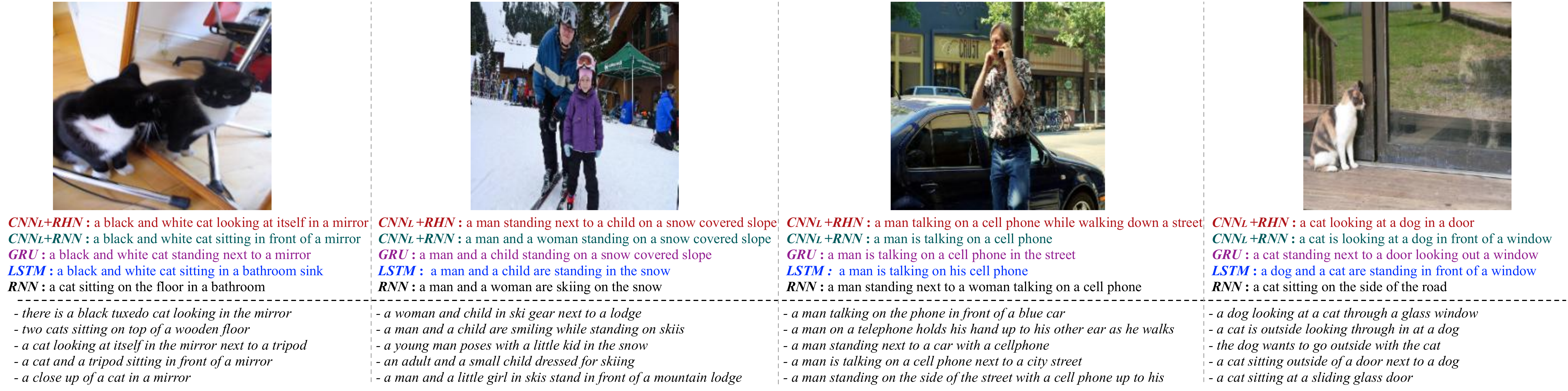}}
		\caption{Qualitative results for images on MS COCO. Ground-truth annotations (under each dashed line) and the generated descriptions are shown for each image.} 
		\label{fig:mscoco_goodresults}
	\end{center}
	\vspace{-5mm}
\end{figure*} 
\begin{figure*}[ht]
	\begin{center}
		\centerline{\includegraphics[width=1.0\textwidth]{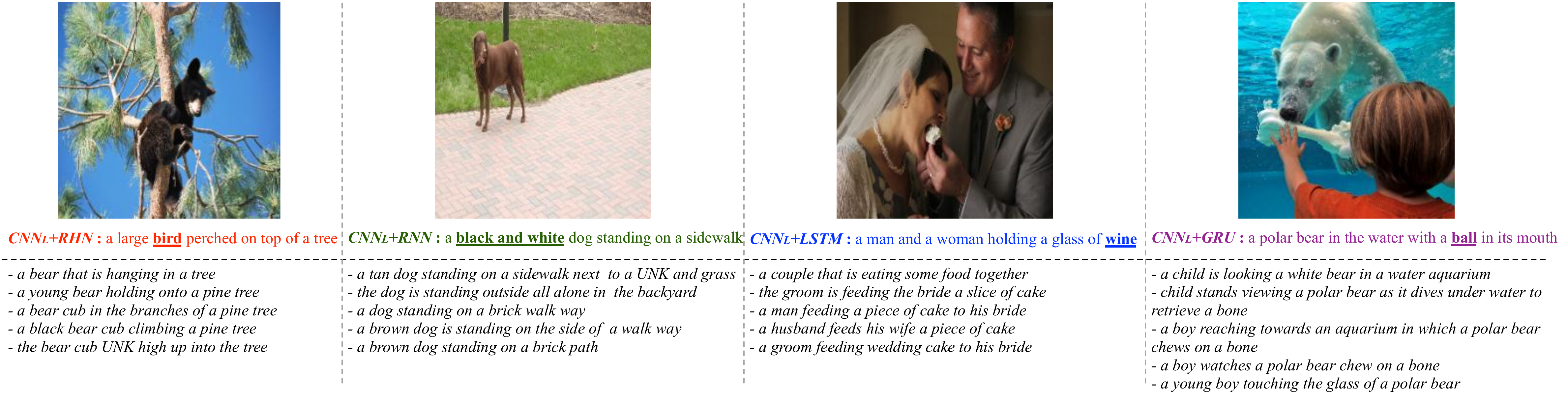}}
		\caption{Some failure descriptions for images on MS COCO. Ground-truth descriptions are under each dashed line.} 
		\label{fig:mscoco_badresults}
	\end{center}
	\vspace{-5mm}
\end{figure*} 

\subsubsection{Comparison with State-of-the-art Methods}
To empirically verify the merit of our models, we compare our methods with other state-of-the-art approaches.

\textbf{Performance on MS COCO.}
The right-hand side of Table~\ref{tab:state_of_the_art} shows the results of different models on MS COCO dataset.
$\text{CNN}_{\mathcal{L}}$-based models perform better than most image captioning models.
The only two methods with better performance (for some metrics) than ours are Attributes-CNN+RNN~\cite{wu2016value} and Google NICv2~\cite{vinyals2016show}.
However, Wu~\etal~\cite{wu2016value} employ an attribute prediction layer, which requires determining an extra attribute vocabulary.
While we generate the image descriptions only based on the image features.
Google NICv2~\cite{vinyals2016show} is based on Google NIC~\cite{vinyals2015show}, the results of Google NICv2 are achieved by model ensembling.
All our models are based on VGG-16 for a fair comparison with~\cite{donahue2015long, fang2015captions, jia2015guiding, mao2014deep, wu2016value, xu2015show}.
Indeed, better image CNN (\eg Resnet~\cite{he2016identity}) leads to higher performance\footnote{We uploaded the results based on Resnet-101+$\text{CNN}_{\mathcal{L}}$+LSTM (named {jxgu\_LCNN\_NTU}) to the official MS COCO evaluation server (\url{https://competitions.codalab.org/competitions/3221}), and achieved competitive ranking across different metrics.}.
Despite all this, the CIDEr score of our $\text{CNN}_{\mathcal{L}}$+LSTM model can still achieve 99.1, which is comparable to their best performance even with a single VGG-16 model.

\textbf{Performance on Flickr30K.}
The results on Flickr30K are reported on the left-hand side of Table~\ref{tab:state_of_the_art}.
Interestingly, $\text{CNN}_{\mathcal{L}}$+RHN performs the best on this smaller dataset and even outperforms the Attributes-CNN+RNN~\cite{wu2016value}.
Obviously, there is a significant performance gap between $\text{CNN}_{\mathcal{L}}$+RNN/RHN/GRU and RNN/RHN/GRU/LSTM models.
This demonstrates the effectiveness of our language CNN on the one hand, and also shows that our $\text{CNN}_{\mathcal{L}}$+RNN/RHN/GRU models are more robust and easier to train than LSTM networks when less training data is available.

\subsection{Qualitative Results}
Figure~\ref{fig:mscoco_goodresults} shows some examples generated by our models.
It is easy to see that all of these caption generation models can generate somewhat relevant sentences, while the $\text{CNN}_{\mathcal{L}}$-based models can predict more high-level words by jointly exploiting history words and image representations.
Take the last image as an example, compared with the sentences generated by RNN/LSTM/GRU model, ``\textit{a cat is looking at a dog in front of a window}" generated by $\text{CNN}_{\mathcal{L}}$+RNN is more precise to describe their relationship in the image. 

Besides, our $\text{CNN}_{\mathcal{L}}$-based models can generate more descriptive sentences.
For instance, with the detected object ``\textit{cat}" in the first image, the generated sentence ``\textit{a black and white cat looking at itself in a mirror}" by $\text{CNN}_{\mathcal{L}}$+RHN depicts the image content more comprehensively.
The results demonstrate that our model with language CNN can generate more humanlike sentences by modeling the hierarchical structure and long-term information of words.

Figure~\ref{fig:mscoco_badresults} shows some failure samples of our $\text{CNN}_{\mathcal{L}}$-based models.
Although most of the generated captions are complete sentences.
However, the biggest problem is that those predicted visual attributes are wrong.
For example, ``\textit{bear}" in the first image is detected as ``\textit{bird}", and ``\textit{brown}" in the second image is detected as ``\textit{black and white}".
This will decrease the precision-based evaluation score (\eg, B@$n$). We can improve our model by further taking high-level attributes into account. 

\section{Conclusion}
In this work, we present an image captioning model with language CNN to explore both hierarchical and temporal information in sequence for image caption generation.
Experiments conducted on MS COCO and Flickr30K image captioning datasets validate our proposal and analysis.
Performance improvements are clearly observed when compared with other image captioning methods.
Future research directions will go towards integrating extra attributes learning into image captioning, and how to apply a single language CNN for image caption generation is worth trying.

\section*{Acknowledgements}
This work is supported by the National Research Foundation, Prime Minister’s Office, Singapore, under its IDM Futures Funding Initiative, and NTU CoE Grant.
This research was carried out at ROSE Lab at Nanyang Technological University, Singapore.
ROSE Lab is supported by the National Research Foundation, Prime Ministers Office, Singapore, under its IDM Futures Funding Initiative and administered by the Interactive and Digital Media Programme Office.
We gratefully acknowledge the support of NVAITC (NVIDIA AI Tech Centre) for our research at NTU ROSE Lab, Singapore.

{\small
\bibliographystyle{ieee}
\bibliography{egbib}
}

\end{document}